\documentclass{article}

\usepackage{microtype}
\usepackage{graphicx}
\usepackage{subfig}
\usepackage{booktabs} 
\usepackage[hyphens]{url}

\usepackage{hyperref}

\newcommand{\ie}{\textit{i.e.}}
\newcommand{\eg}{\textit{e.g.}}

\usepackage{colortbl}
\usepackage{arydshln}
\usepackage{enumitem}
\usepackage{multirow}
\usepackage[most]{tcolorbox}

\definecolor{mygray}{RGB}{226, 226, 226}
\definecolor{myred}{RGB}{252, 142, 142}
\definecolor{mygreen}{RGB}{147, 255, 143}
\definecolor{myblue}{RGB}{144, 155, 255}
\definecolor{myyellow}{RGB}{253, 253, 143}
\definecolor{mypurple}{RGB}{255, 142, 250}

\tcbset{
  aibox/.style={
    top=8pt,
    bottom=3pt,
    colback=white,
    colframe=black,
    colbacktitle=black,
    enhanced,
    center,
    attach boxed title to top left={yshift=-0.1in,xshift=0.15in},
    boxed title style={boxrule=0pt,colframe=white,},
  }
}
\newtcolorbox{AIbox}[3][]{aibox, width=#2, title=#3,#1}

\usepackage[preprint]{icml}

\usepackage{amsmath}
\usepackage{amssymb}
\usepackage{mathtools}
\usepackage{amsthm}

\usepackage[capitalize,noabbrev]{cleveref}

\theoremstyle{plain}
\newtheorem{theorem}{Theorem}[section]
\newtheorem{proposition}[theorem]{Proposition}

\newtheorem{corollary}[theorem]{Corollary}
\theoremstyle{definition}

\theoremstyle{remark}

\usepackage[textsize=tiny]{todonotes}

\begin{document}

\twocolumn[
    \icmltitle{On the Failure of Latent State Persistence in Large Language Models}
    
    
    
    \icmlsetsymbol{equal}{*}
    
    \begin{icmlauthorlist}
        \icmlauthor{Jen-tse Huang}{jhu}
        \icmlauthor{Kaiser Sun}{jhu}
        \icmlauthor{Wenxuan Wang}{ruc}
        \icmlauthor{Mark Dredze}{jhu}
    \end{icmlauthorlist}
    
    \icmlaffiliation{jhu}{Johns Hopkins University}
    \icmlaffiliation{ruc}{Renmin University of China}
    
    \icmlcorrespondingauthor{Jen-tse Huang}{jhuan236@jh.edu}
    
    \icmlkeywords{Machine Learning, ICML}
    
    \vskip 0.3in
]



\printAffiliationsAndNotice{}  

\begin{abstract}

While Large Language Models (LLMs) excel in reasoning, whether they can sustain persistent latent states remains under-explored.
The capacity to maintain and manipulate unexpressed, internal representations—analogous to human working memory—is a cornerstone of complex reasoning.
In this paper, we formalize and quantify the ``Latent State Persistence'' (LSP) gap through three novel experiments.
First, we utilize a Number Guessing Game, demonstrating that across independent queries, LLMs fail to allocate probability mass to a singular hidden choice, violating a fundamental probabilistic principle.
Second, we employ a Yes-No Game to show that as the number of questions increases, LLMs suffer from ``concept drift,'' leading to inevitable self-contradictions due to the lack of LSP.
Finally, inspired by Mathematical Mentalism, we task models with tracking transformations on hidden variables, revealing a failure in variable binding and state evolution when the initial state is not explicitly present in the context.
Collectively, these findings suggest that LLMs function as reactive post-hoc solvers rather than proactive planners with LSP.
Our work provides a framework for evaluating the fidelity of internal representations and highlights a fundamental architectural divergence between autoregressive transformers and human-like cognition.
\end{abstract}

\section{Introduction}
\label{sec:intro}

Large Language Models (LLMs) have demonstrated remarkable proficiency across diverse reasoning domains, from legal analysis and education to complex healthcare applications~\cite{guha2023legalbench, wen2024ai, huang2025competing, jiao2023chatgpt, yang2024talk2care}.
Consequently, research has increasingly focused on whether these models exhibit human-like cognitive traits such as personality, empathy, and theory-of-mind~\cite{huang2024humanity, huang2024apathetic, sorin2024large, liu2024interintent, wang2024not}.
However, a fundamental pillar of cognition remains significantly under-explored: working memory—the capacity to temporally store and manipulate internal representations without externalizing them.

For Transformers, the ``state'' of a conversation is traditionally bounded by the context window.
This raises a critical architectural question: can an LLM maintain a persistent latent state that is not explicitly written into the prompt?
We formalize this capability as Latent State Persistence (LSP)—the ability of a model to instantiate, maintain, and manipulate an internal variable $x$ across multiple interaction steps without explicit context externalization.
We hypothesize that current LLMs lack true LSP, functioning as reactive post-hoc reasoners that find the most statistically plausible completion for a given dialogue pattern rather than grounding their responses in a stable internal state.

The central challenge in evaluating LSP is the ``hidden mind'' problem: how can we verify the existence of an unobservable internal state?
We address this by proposing three novel experiments, each deriving a testable hypothesis from the mathematical consequences of LSP:
\begin{itemize}[noitemsep]
    \item \textbf{Number Guessing Game (\S\ref{sec:number-guessing})} tests \textit{Static Probabilistic Completeness}: We propose the Sum-of-Probability Identity (Proposition \ref{prop:sum_prob}) as a null hypothesis. For a state-persistent agent, the sum of ``Yes'' probabilities across a complete state space must equal unity ($\sum P = 1$). We define Empirical State Mass (ESM) as a metric to quantify the deviation from this identity.
    \item \textbf{Yes-No Game (\S\ref{sec:yes-no})} examines \textit{Dynamic Logical Consistency}: We demonstrate that for an agent with LSP, inference complexity is constant-time $O(1)$ relative to the number of queries (Proposition \ref{prop:o1_complexity}). Conversely, a stateless agent must treat the task with $O(t)$ complexity. We show that as constraints accumulate, models inevitably fail due to ``concept drift'' and self-contradiction.
    \item \textbf{Mathematical Mentalism (\S\ref{sec:math-magic})} assesses \textit{State Evolution Fidelity}: Inspired by deterministic magic routines, we task models with tracking non-linear transformations on hidden variables. Failure to satisfy the mathematical invariants of the routine reveals a breakdown in variable binding and latent state evolution.
\end{itemize}

We evaluate 17 frontier LLMs, including GPT (4o~\cite{gpt4o} and 4o-Mini~\cite{gpt4omini}), o-series (o1-Mini~\cite{o1}, o3-Mini, and o4-Mini~\cite{o3-o4mini}), LLaMA (3.1~\cite{llama31} and 3.3), Qwen-2.5~\cite{qwen25} (7B and 72B), QwQ~\cite{qwq}, and DeepSeek (V3~\cite{deepseekv3} and R1~\cite{deepseekr1}).
Temperature and Top‑$p$ are default to $1.0$.
Our results converge on a singular conclusion: current LLMs exhibit a profound ``LSP gap.''
While scaling and long-reasoning traces (\eg, in o3-Mini or R1) improve performance by externalizing reasoning, the underlying latent persistence remains absent.

Our framework distinguishes itself from existing ``working memory'' benchmarks, such as N-back tasks \cite{gong2024working, zhang2024working}, which often conflate internal maintenance with context-window retrieval (Appendix \ref{sec:working-memory-eval}).
By ensuring the target state is never present in the input, our methodology provides a faithful evaluation of true working memory.
Our contributions are as follows:
\begin{enumerate}[noitemsep]
    \item We formally define \textbf{Latent State Persistence (LSP)} and provide a theoretical framework to detect hidden states via probabilistic and logical indicators.
    \item We introduce our framework\footnote{\url{https://github.com/penguinnnnn/LLM-Working-Memory}} containing three experiments designed to quantify the fidelity of internal representations in autoregressive models. 
    \item We extensively evaluate 17 frontier LLMs, revealing that scaling laws fail to bridge the LSP gap and documenting the persistence of human-centric priors, such as the ``Blue-Seven'' phenomenon.
\end{enumerate}
\section{Number Guessing Game}
\label{sec:number-guessing}

\subsection{Methodology}

\paragraph{Problem Formalization.}
Let $\mathcal{X} = \{1, \dots, n\}$ be a discrete state space of integers.
We define an experimental trial as a sequence of interactions $\mathcal{T}_i = (P_{\text{inst}}, R_{\text{ack}}, Q_i)$, where:
\begin{itemize}[noitemsep]
    \item $P_{\text{inst}}$ is the initial instruction: ``Think of an integer between 1 and $n$, but don't say it to me.''
    \item $R_{\text{ack}}$ is the model's acknowledgement: ``Got it! I've thought of an integer between 1 and $n$. What's next?''
    \item $Q_i$ is the query: ``Is the number you're thinking of $i$? Answer Yes or No.''
\end{itemize}

\paragraph{The Persistence Hypothesis.}
If an agent possesses perfect LSP, it must instantiate and maintain a fixed latent variable $x \in \mathcal{X}$ throughout the trial $\mathcal{T}_i$.
Under another assumption of honesty (the agent does not lie), the response $y \in \{\text{Y}, \text{N}\}$ distribution is conditioned on this hidden state $x$:
\begin{equation}
    P(y = \text{Y} \mid Q_i, x) = \mathbb{I}(i = x),
\end{equation}
where $\mathbb{I}(\cdot)$ is the indicator function.
\begin{proposition}[Sum-of-Probability Identity]
\label{prop:sum_prob}
For an agent with a persistent and unique latent state $x \perp Q_i$, the sum of the probabilities of answering ``Yes'' across all possible queries in $\mathcal{X}$ must equal unity:
\begin{equation}
\begin{aligned}
    \sum_{i=1}^{n} P(y = \text{Y} \mid Q_i) &= \sum_{i=1}^{n} \sum_{x \in \mathcal{X}} P(y = \text{Y} \mid Q_i, x) P(x) \\
    &= \sum_{x \in \mathcal{X}} P(x) \sum_{i=1}^{n} \mathbb{I}(i = x) = 1.
\end{aligned}
\end{equation}
\end{proposition}
Any significant deviation from $\sum P = 1$ provides a quantitative basis to reject the hypothesis, suggesting the agent either fails to commit to a state $x$ or lies about the outcome.

\paragraph{LLMs as Stateless Agents.}
To quantify this phenomenon in stateless agents, we define the Empirical State Mass (ESM) for a model $M$ over $\mathcal{X}$ as:
\begin{equation}
    \text{ESM}(M, \mathcal{X}) = \sum_{i=1}^{n} \hat{P}_M(y = \text{Y} \mid \mathcal{T}_i),
\end{equation}
where $\hat{P}_M$ is the empirical probability derived from repeated independent samplings.
In a stateless agent, the model does not hold a fixed $x$.
Instead, it treats each $Q_i$ as a frequentist inference task, leading to $x \not\perp Q_i$.
Since $P(i=x)$ is a low-probability event (\eg, $0.1$ for $n=10$), an autoregressive model biased toward maximum likelihood estimation (MLE) will favor the majority class, leading to $\text{ESM} \to 0$.

\paragraph{Experimental Setup and Metrics.}
We evaluated 17 frontier LLMs using different $n$.
To ensure statistical significance, each query $Q_i$ was executed for 200 independent trials.
We varied the decoding configurations, Temperature and Top-$p$, ranging from 0 to 1, to test the transition from greedy decoding to stochastic sampling and the sensitivity of ESM to the probability mass of the tail tokens.

\subsection{Results}

\begin{figure*}[t]
  \centering
  \subfloat[OpenAI GPT models.]{
    \includegraphics[width=0.49\linewidth]{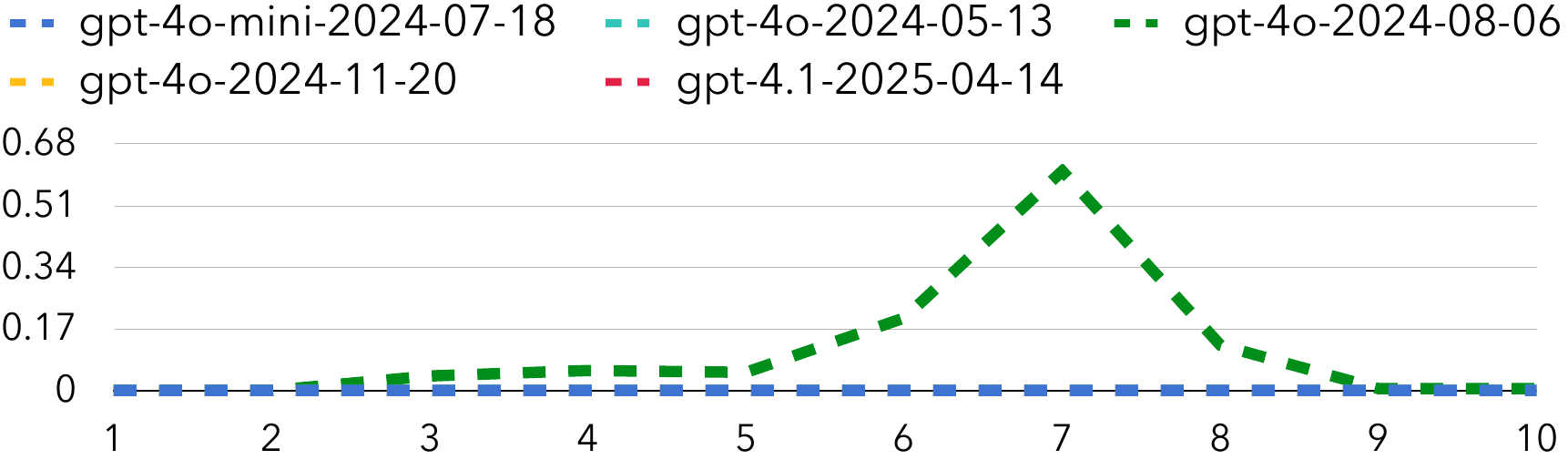}
    \label{fig:number-gpt}
  }
  \subfloat[OpenAI o-series models.]{
    \includegraphics[width=0.49\linewidth]{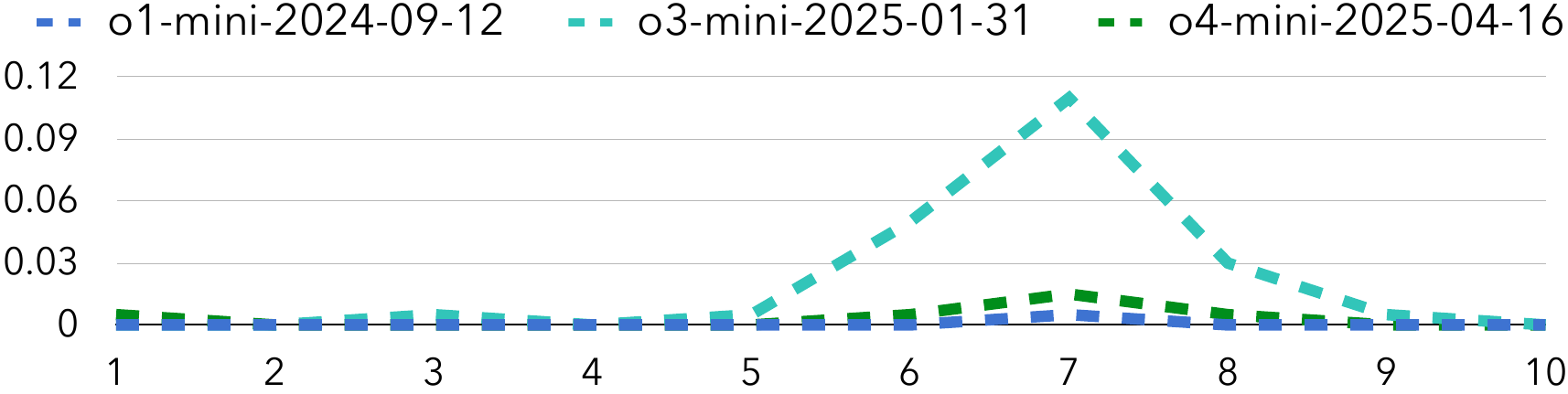}
    \label{fig:nuber-o1}
  } \\
  \subfloat[Meta LLaMA models.]{
    \includegraphics[width=0.49\linewidth]{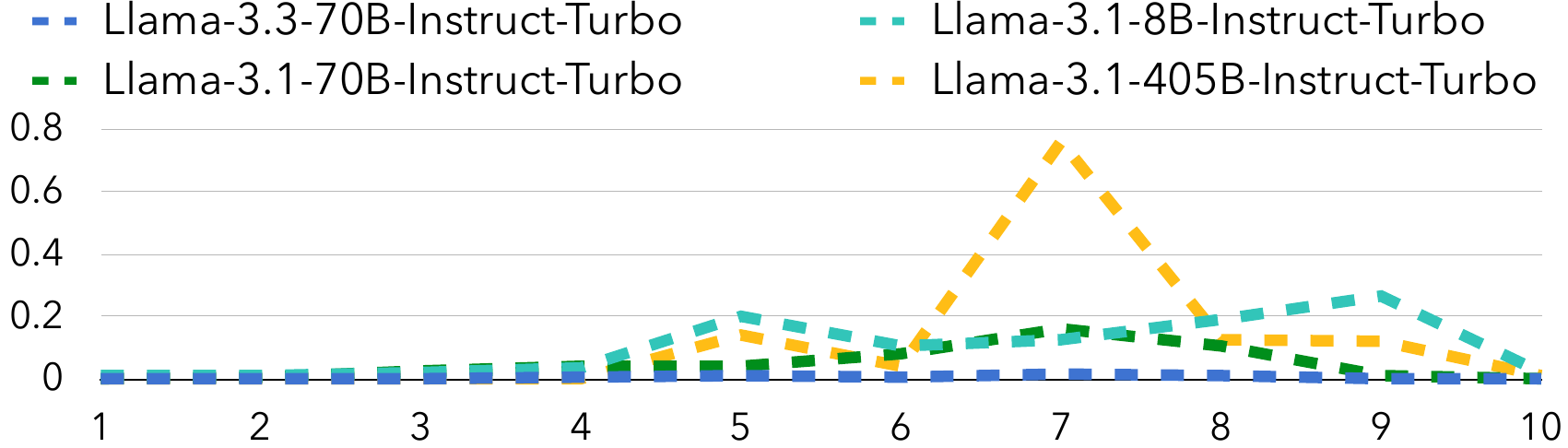}
    \label{fig:number-llama}
  }
  \subfloat[Qwen and DeepSeek models.]{
    \includegraphics[width=0.49\linewidth]{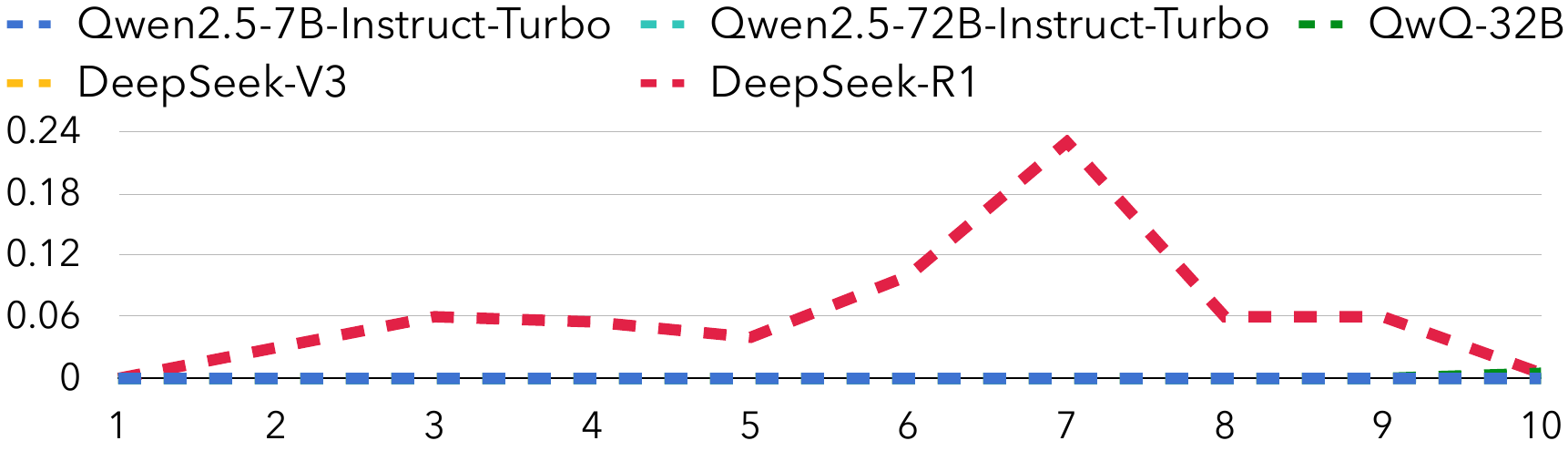}
    \label{fig:number-qwen-deepseek}
  }
  \caption{Empirical distribution of affirmative responses for $i \in \{1, \dots, 10\}$. A pronounced horizontal band at $i=7$ reveals a pervasive ``blue-seven'' heuristic bias across most models, regardless of their scale or architecture.}
  \label{fig:number}
\end{figure*}

\paragraph{Qualitative Analysis.}
Figure~\ref{fig:number} illustrates the empirical distribution of ``Yes'' responses for $n=10$.
Two primary phenomena emerge.
(1) \textbf{Systematic under-allocation of probability mass}: the majority of tested LLMs exhibit a severe negative bias, where ``No'' responses dominate regardless of $i$.
This leads to an ESM approaching zero, violating Proposition~\ref{prop:sum_prob}.
We posit that the models are not querying an internal state but are instead performing frequentist inference on the likelihood of a human guess being correct ($1/n$), which is statistically low.
(2) When models do provide affirmative responses, they exhibit a pronounced heuristic bias toward the number seven, echoing the ``blue-seven'' phenomenon observed in human psychology~\cite{miller1956magical, kubovy1976predominance}.

\begin{table}[t]
\centering
\caption{ESM across 17 evaluated LLMs ($n=10$). Color intensity denotes the deviation from the theoretical identity $\text{ESM}=1$: red indicates under-allocation, while blue indicates over-allocation.}
\label{tab:prob-sum}
\begin{tabular}{ll}
\toprule
\textbf{Model} & \textbf{ESM} \\
\midrule
GPT-4o-Mini-2024-07-18 & \cellcolor{myred!100} 0 \\
GPT-4o-2024-05-13 & \cellcolor{myred!100} 0 \\
GPT-4o-2024-08-06 & \cellcolor{myblue!8.5} 1.085 \\
GPT-4o-2024-11-20 & \cellcolor{myred!100} 0 \\
GPT-4.1-2025-04-14 & \cellcolor{myred!100} 0 \\
\hdashline
o1-Mini-2024-09-12 & \cellcolor{myred!99.5} 0.005 \\
o3-Mini-2025-01-31 & \cellcolor{myred!79.5} 0.205 \\
o4-Mini-2025-04-16 & \cellcolor{myred!97} 0.030 \\
\hdashline
LLaMA-3.3-70B-Instruct-Turbo & \cellcolor{myred!95.5} 0.045 \\
LLaMA-3.1-8B-Instruct-Turbo & \cellcolor{myred!2} 0.980 \\
LLaMA-3.1-70B-Instruct-Turbo & \cellcolor{myred!53.5} 0.465 \\
LLaMA-3.1-405B-Instruct-Turbo & \cellcolor{myblue!19.5} 1.195 \\
\hdashline
Qwen2.5-7B-Instruct-Turbo & \cellcolor{myred!100} 0 \\
Qwen2.5-72B-Instruct-Turbo & \cellcolor{myred!100} 0 \\
QwQ-32B & \cellcolor{myred!99.5} 0.005 \\
\hdashline
DeepSeek-V3 & \cellcolor{myred!100} 0 \\
DeepSeek-R1 & \cellcolor{myred!36} 0.640 \\
\bottomrule
\end{tabular}
\end{table}

\paragraph{Quantitative Analysis.}
Table~\ref{tab:prob-sum} reports the ESM across all models.
We highlight three critical observations:
(1) \textbf{Non-monotonic Scaling:} Model performance does not scale linearly with size or recency.
In the GPT family, GPT-4o-0806 yields the highest ESM, outperforming both its successor (GPT-4o-1120) and GPT-4.
Notably, the ESM of LLaMA-3.1-8B is closer to one, compared to its larger 70B and 405B counterparts.
(2) \textbf{Invariance to Reasoning Traces:} Contrary to expectations, models featuring internal CoT—such as o1, o3, and DeepSeek-R1—do not bridge the LSP gap.
Despite extended compute, these models still fail to instantiate a continuous and consistent LSP, treating the task as a reactive prompt-completion exercise.
(3) \textbf{Stochasticity of Capability:} The acquisition of this state-like behavior appears stochastic rather than a predictable emergent property of scale, suggesting that current training objectives do not explicitly incentivize latent state maintenance.

\begin{figure*}[h]
    \centering
    \includegraphics[width=1.0\linewidth]{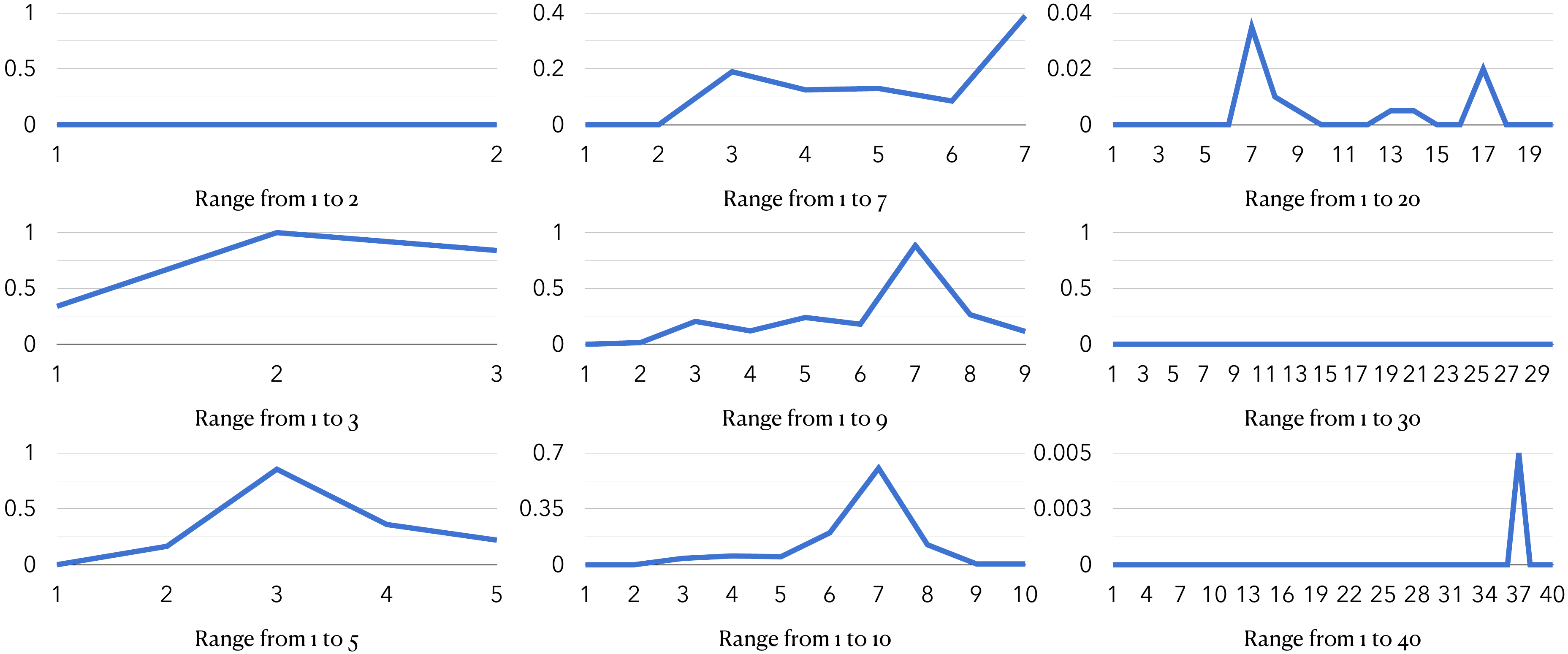}
    \caption{Empirical distributions of affirmative responses for GPT-4o-2024-08-06 across varying state space sizes.}
    \label{fig:number-range}
\end{figure*}

\paragraph{Sensitivity to State Space Size.}
We extended the experiment to varying ranges $n \in \{3, \dots, 40\}$ using GPT-4o-2024-08-06.
As shown in Table~\ref{tab:prob-sum-range} and Figure~\ref{fig:number-range}, the model's ESM is highly sensitive to the cardinality of $\mathcal{X}$.
For small ranges ($n < 10$), the model becomes over-optimistic, yielding an ESM significantly greater than 1.
For larger ranges ($n \geq 20$), the ESM collapses toward zero.
Furthermore, the bias toward numbers ending in seven (\eg, 7, 17, 37) persists across all ranges, suggesting that the model relies on digit-level heuristics rather than a uniform sampling of the state space.

\begin{table}[t]
\centering
\caption{Sensitivity of ESM to state space cardinality ($n$) for GPT-4o-2024-08-06. The decay suggests the model's threshold is coupled with the prior probability $1/n$ rather than a LSP.}
\label{tab:prob-sum-range}
\resizebox{1.0\linewidth}{!}{
\setlength{\tabcolsep}{3pt}
\begin{tabular}{lccccccccc}
\toprule
\bf Number & 2 & 3 & 5 & 7 & 9 & 10 & 20 & 30 & 40 \\
\midrule
\bf ESM & \cellcolor{myred!100} 0 & \cellcolor{myblue!118} 2.18 & \cellcolor{myblue!60} 1.6 & \cellcolor{myred!8} 0.92 & \cellcolor{myblue!102.5} 2.025 & \cellcolor{myblue!8.5} 1.085 & \cellcolor{myred!92} 0.08 & \cellcolor{myred!100} 0 & \cellcolor{myred!99.5} 0.005 \\
\bottomrule
\end{tabular}
}
\end{table}

\paragraph{Ablation of Decoding Strategies.}
To verify if these failures are artifacts of decoding, we evaluated the ESM under varying Temperature ($T$) and Top-$p$ settings with GPT-4o-2024-08-06.
As shown in Table~\ref{tab:decoding}, while higher $T$ increases the variance of responses and lower Top-$p$ concentrates more mass on the number 7, the qualitative conclusion does not change: the ESM of this model never converges to zero as other models show.
This confirms that the capacity to maintain internal variables outside the explicit context is not related to surface-level decoding but a more fundamental architectural issue.

\paragraph{Implication.}
The collective evidence from the Number Guessing Game suggests that LLMs operate as reactive post-hoc reasoners.
When an LLM claims to have ``thought of a number,'' it does not commit to a latent $x \in \mathcal{X}$.
Instead, it generates a response based on the statistical likelihood of truthiness within the given dialogue pattern.
This lack of LSP results in a breakdown of the most basic probabilistic requirement for a hidden-state agent that the sum must equal one, manifesting core challenges in AI alignment—sycophancy and false promises.
\section{Yes-No Game}
\label{sec:yes-no}

\subsection{Methodology}

\paragraph{Problem Formalization.}
The ``Yes-No'' game (a variant of \textit{Twenty Questions}\footnote{\url{https://en.wikipedia.org/wiki/Twenty_questions}}) is a classic social deduction task requiring persistent reasoning and categorical narrowing.
In our setting, an agent privately selects an object, and the opponent poses a sequence of binary queries $Q$ to identify the object.
We define a multi-dimensional object space $\mathcal{X}$ characterized by a set of $k$ totally ordered attributes $\mathcal{A} = \{a_1, \dots, a_k\}$.
For each attribute $a \in \mathcal{A}$, there exists a scalar mapping $f_a: \mathcal{X} \to \mathbb{R}$, enabling a total ordering of objects.
A trial $\mathcal{T}$ is defined as a sequence: $\mathcal{T} = (P_{\text{inst}}, R_{\text{ack}}, (Q_1, y_1), \dots, (Q_t, y_t))$, where $P_{\text{inst}}$ and $R_{\text{ack}}$ share the same definition with the Number Guessing Game, $Q_t$ is a query of the form ``Is $f_a(x) \text{ op } f_a(x_{\text{ref}})$?'' and $y_t \in \{\text{Y, N}\}$ is the agent's response.

\begin{table}[t]
    \centering
    \caption{Impact of decoding configurations, Temperature (T) and Top-$p$ (P), on ESM for GPT-4o-2024-08-06 with $n=10$.}
    \label{tab:decoding}
    \resizebox{1.0\linewidth}{!}{
    \setlength{\tabcolsep}{3pt}
    \begin{tabular}{lcccccccccc|c}
    \toprule
    \bf Settings & \bf 1 & \bf 2 & \bf 3 & \bf 4 & \bf 5 & \bf 6 & \bf 7 & \bf 8 & \bf 9 & \bf 10 & \bf ESM \\
    \midrule
    $T_{1.0} P_{1.0}$ & 0 & 0 & 0.04 & 0.055 & 0.05 & 0.2 & 0.605 & 0.125 & 0.005 & 0.005 & 1.085 \\
    $T_{0.7} P_{1.0}$ & 0 & 0.005 & 0.01 & 0.09 & 0.055 & 0.225 & 0.765 & 0.09 & 0.005 & 0 & 1.245 \\
    $T_{0.4} P_{1.0}$ & 0 & 0 & 0 & 0.025 & 0.005 & 0.165 & 0.86 & 0.04 & 0 & 0 & 1.095 \\
    $T_{0.1} P_{1.0}$ & 0 & 0 & 0 & 0 & 0 & 0.085 & 0.895 & 0.005 & 0 & 0 & 0.985 \\
    $T_{1.0} P_{0.7}$ & 0 & 0 & 0 & 0 & 0 & 0.1 & 0.83 & 0.02 & 0 & 0 & 0.95 \\
    $T_{1.0} P_{0.4}$ & 0 & 0 & 0 & 0 & 0 & 0.03 & 0.825 & 0 & 0 & 0 & 0.855 \\
    $T_{1.0} P_{0.1}$ & 0 & 0 & 0 & 0 & 0 & 0.02 & 0.82 & 0 & 0 & 0 & 0.84 \\
    \bottomrule
    \end{tabular}
    }
\end{table}

\paragraph{The Persistence Hypothesis.}
An agent exhibiting perfect LSP must instantiate and maintain a fixed latent variable $x$ throughout the trial $\mathcal{T}$.
For any query $Q_t$ involving a reference object $x_\text{ref} \in \mathcal{X}$, an attribute $a$, and a comparison operator $\text{op} \in \{>, <\}$, the response $y_t$ is determined by:
\begin{equation}
    y_t = \mathbb{I}(f_a(x) \text{ op } f_a(x_\text{ref})).
\end{equation}
\begin{proposition}[Constant-Time Inference]
\label{prop:o1_complexity}
For an agent with a persistent latent state $x$, the computational complexity of generating $y_t$ is $O(1)$ relative to the trial length $t$, as it requires only a singular comparison between $x$ and $x_{\text{ref}}$.
\end{proposition}
Crucially, human players do not typically perform exhaustive cross-checks against the history of $t-1$ responses; instead, they rely on the LSP to ensure global consistency.

\paragraph{LLMs as Post-Hoc Constraint Solvers.}
In contrast, an agent lacking LSP (\eg, an autoregressive LLM) must determine $y_t$ solely based on the conditional probability $P_M(y_t \mid \mathcal{T}_{<t})$.
Without a grounded latent variable $x$, the model is tasked with solving a Dynamic Constraint Satisfaction Problem (DCSP).
To remain consistent, the model must ensure that $y_t$ satisfies the intersection of all prior constraints $\{(Q_\tau, y_\tau)\}_{\tau=1}^{t-1}$.
As $t$ increases, the constraint density grows, requiring the model to navigate an increasingly complex logical manifold within its context window.
This shifts the task from simple retrieval to complex $O(t)$ (or higher) relational reasoning.
We hypothesize that LLMs, acting as reactive post-hoc reasoners, will inevitably suffer from ``concept drift'' and violate earlier constraints once the combinatorial complexity exceeds their inherent reasoning capacity.
For instance, the model might initially answer ``Yes'' to ``Is the object heavier than an elephant?'' but later also respond ``Yes'' to ``Is the object lighter than a cat?'', thereby contradicting itself.

\begin{table}[t]
\centering
\caption{Ground-truth for the objects sorted across five physical attributes: volume, length, weight, density, and hardness.}
\label{tab:object-list}
\resizebox{1.0\linewidth}{!}{
\begin{tabular}{llll}
\toprule
\multicolumn{1}{c}{\colorbox{mygray}{\textbf{Volume}}} & \multicolumn{1}{c}{\colorbox{mygray}{\textbf{Length}}} & \multicolumn{1}{c}{\colorbox{mygray}{\textbf{Weight}}} & \multicolumn{1}{c}{\colorbox{mygray}{\textbf{Density}}} \\
Coffee bean & Rice & Coin & Air \\
Dice & Paperclip & Spoon & Wood \\
Golf ball & Credit card & Watch & Ice \\
Soda can & Pencil & Smartphone & Water \\
Soccer ball & Laptop & Bottle of water & Plastic \\
Microwave oven & Baseball bat & Dictionary & Glass \\
Washing machine & Guitar & Cat & Iron \\
Bathtub & Door & Bicycle & Copper \\
Car & Apple tree & Television & Silver \\
School bus & Coconut tree & Refrigerator & Gold \\
Shipping container & Tennis court & Tiger & \multicolumn{1}{c}{\colorbox{mygray}{\textbf{Hardness}}} \\
Olympic swimming pool & Swimming pool & Cow & Marshmallow \\
Boeing 747 & Football field & Rhino & Rubber eraser \\
Titanic & Skyscraper & Elephant & Brick \\
Great Pyramid of Giza & Mount Everest & Train & Hammer \\
& & & Diamond ring \\
\bottomrule
\end{tabular}
}
\end{table}

\paragraph{Experimental Setup and Metrics.}
We curate a space $\mathcal{X}$ consisting of 60 distinct objects across five physical dimensions: volume, length, weight, density, and hardness.
The ground-truth ordering for each attribute is established a priori (see Table~\ref{tab:object-list}).
In each trial, queries are generated by randomly sampling an attribute $a \in \mathcal{A}$, a reference object $x_{\text{ref}} \in \mathcal{X}$, and an operator (\eg, bigger or smaller for volume).
We cap each trial at $T=250$ queries.
To quantify self-contradiction, we define the Feasible Set $\mathcal{S}_t$ at step $t$:
\begin{equation}
    \mathcal{S}_t = \{x \in \mathcal{X} \mid \forall \tau \leq t, \text{is\_consistent}(x, Q_\tau, y_\tau)\}.
\end{equation}
\begin{corollary}[Self-Consistency wish LSP]
\label{coro:self_consistency}
The ability of an agent with perfect LSP to keep all responses logically consistent, \ie, $\mathcal{S}_t \neq \emptyset$, is unrelated to $t$.
\end{corollary}
A logical failure occurs at the first $t$ where $\mathcal{S}_t = \emptyset$, indicating that no object in $\mathcal{X}$ can satisfy the accumulated constraints.
We report the Mean Steps to Contradiction (MSC) and the Pass Rate (PR, the percentage of trials reaching $t=250$ without contradiction) across 200 independent trials for GPT-4o-Mini-0718 and GPT-4o-0806.

\begin{table}[t]
    \centering
    \caption{Performance comparison between GPT-4o and GPT-4o-Mini using MSC, PR, and failure count with $T=250$.}
    \label{tab:object-result}
    \resizebox{1.0\linewidth}{!}{
    \begin{tabular}{lcc|cccccc}
    \toprule
    \bf Model & \bf MSC & \bf PR & \bf Failure & \bf V & \bf W & \bf L & \bf D & \bf H \\
    \midrule
    GPT-4o-Mini & 41.43 & 0.0 & 200 & 12 & 46 & 49 & 52 & 41 \\
    GPT-4o & 74.55 & 13.5 & 173 & 21 & 42 & 57 & 27 & 26 \\
    \bottomrule
    \end{tabular}
    }
\end{table}

\begin{table}[t]
    \centering
    \caption{Ablation of knowledge vs. grounding using GPT-4o.}
    \label{tab:ablation}
    \resizebox{1.0\linewidth}{!}{
    \begin{tabular}{lcc|cccccc}
    \toprule
    \bf Model & \bf MSC & \bf PR & \bf Failure & \bf V & \bf W & \bf L & \bf D & \bf H \\
    \midrule
    Hint & 55.34 & 3.0 & 194 & 37 & 39 & 60 & 37 & 21 \\
    All & 68.33 & 21.0 & 158 & 18 & 29 & 21 & 55 & 35 \\
    Hint + All & 87.81 & 27.5 & 145 & 13 & 32 & 34 & 46 & 20 \\
    \bottomrule
    \end{tabular}
    }
\end{table}

\subsection{Results}

\begin{figure}[t]
  \centering
  \subfloat[GPT-4o-Mini-2024-07-18.]{
    \includegraphics[width=1.0\linewidth]{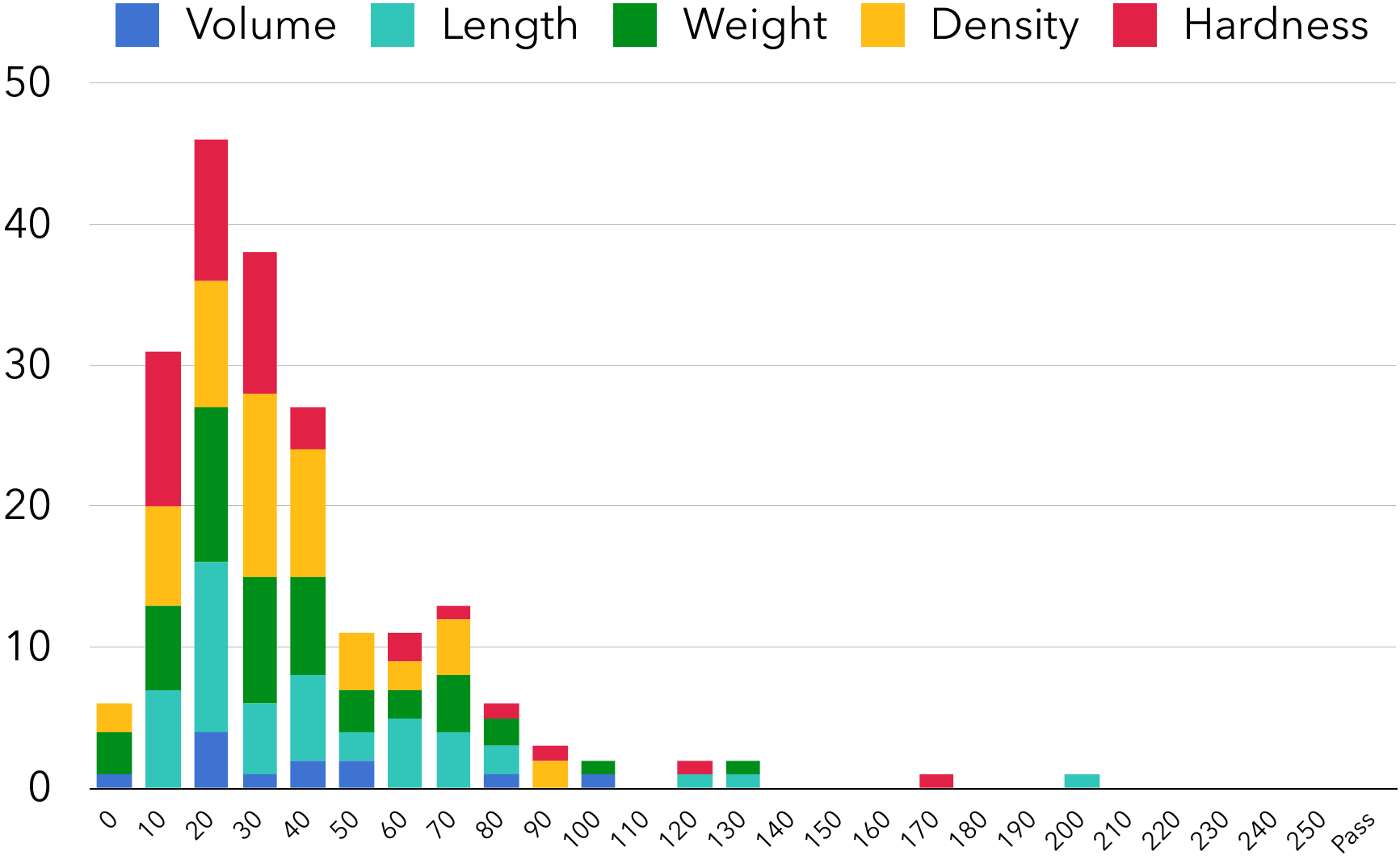}
    \label{fig:object-gpt-4o-mini}
  }
  \\
  \subfloat[GPT-4o-2024-08-06.]{
    \includegraphics[width=1.0\linewidth]{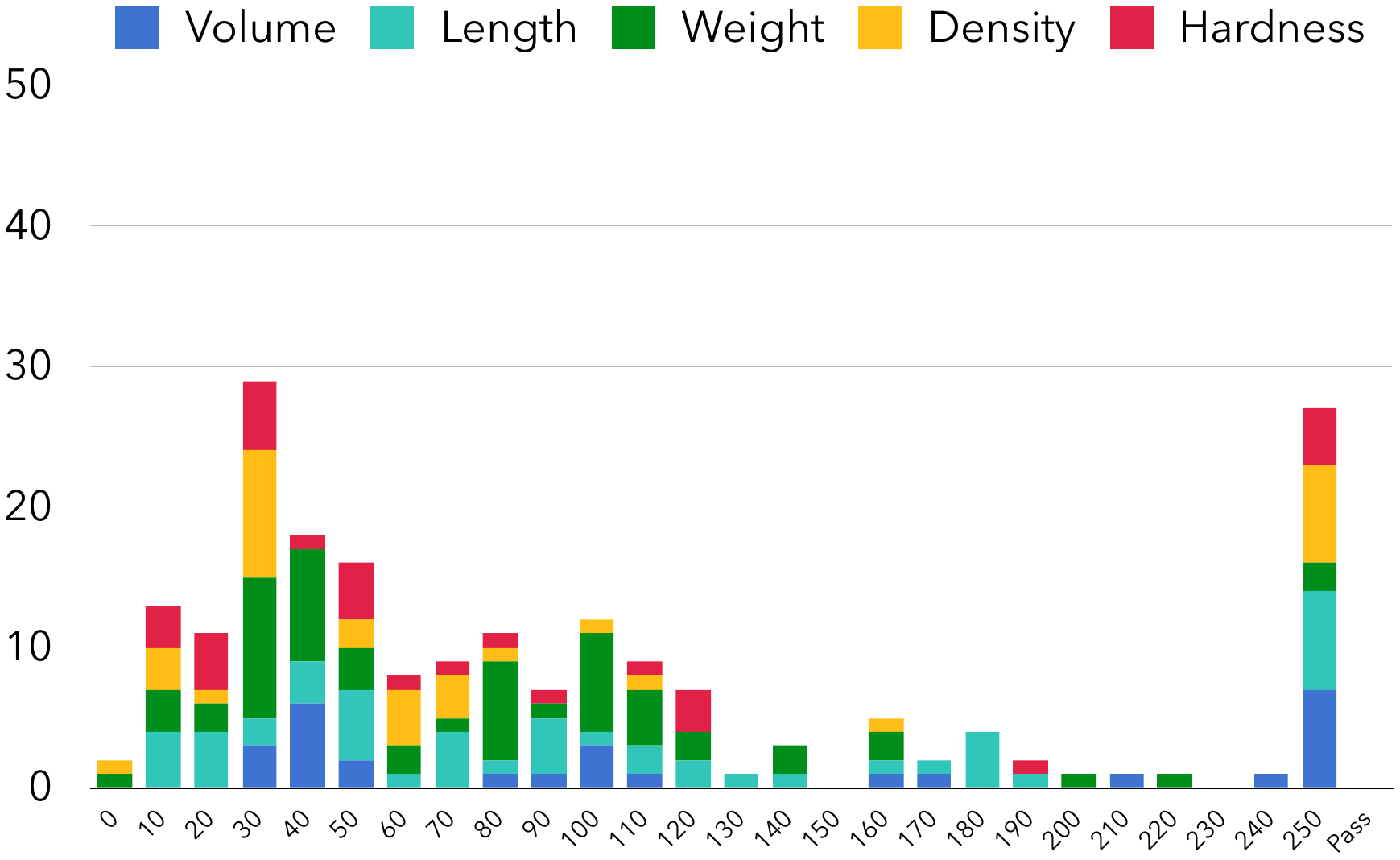}
    \label{fig:object-gpt-4o}
  }
  \caption{Distribution of steps to logical contradiction. The histograms illustrate the trial lengths before $\mathcal{S}_t = \emptyset$.}
  \label{fig:object}
  \vspace{-8pt}
\end{figure}

\paragraph{Qualitative Analysis.}
Figure~\ref{fig:object} characterizes the distribution of trial lengths before a logical contradiction occurs.
For GPT-4o-Mini, the failure distribution is highly concentrated, peaking within the 20--30 query range.
In contrast, GPT-4o exhibits a broader distribution with a peak shifted to the 30--40 range and a significantly thicker tail in the 80--130 range, suggesting superior but still finite constraint-handling capabilities.
Attribute-wise analysis reveals that GPT-4o-Mini is particularly vulnerable to attributes requiring abstract physical reasoning, such as density and hardness, whereas GPT-4o demonstrates relative robustness on these dimensions but eventually fails due to the accumulated combinatorial complexity.

\paragraph{Quantitative Analysis.}
Table~\ref{tab:object-result} summarizes the MSC and PR.
The disparity between the two models is obvious: GPT-4o-Mini fails to complete a single trial without contradiction (PR=0\%, MSC=41.43), whereas GPT-4o achieves a PR of 13.5\% with a substantially higher MSC of 74.55.
These results support our hypothesis that performance in the Yes-No Game scales with the model's capacity for long-context relational reasoning, indicating a linear $O(t)$ inference complexity instead of constant $O(1)$.
Crucially, the non-zero PR of GPT-4o does not necessarily imply the presence of LSP; rather, it reflects a more sophisticated ability to maintain a valid intersection of constraints within the context window for a longer duration.

\paragraph{Ablation Studies: Knowledge vs. Grounding.}
To decouple the failure of LSP from potential deficits in world knowledge (\ie, the model's inability to rank objects), we conduct three ablation settings (Table~\ref{tab:ablation}):
\begin{itemize}[noitemsep]
    \item \textbf{(1) Hint}: Providing the ground-truth attribute rankings (defined in Table~\ref{tab:object-list}) in the system prompt.
    \item \textbf{(2) All}: Explicitly specifying the target object to the model at each step, thereby externalizing the $x$.
    \item \textbf{(3) Hint + All}: Combining both external knowledge and state grounding.
\end{itemize}
Our findings indicate that \textbf{Hint} even decreases the performance, confirming that contradictions stem from logical drift rather than factual ignorance.
Conversely, \textbf{Hint + All} substantially elevates MSC and PR, as it collapses the DCSP into a series of independent, short-context comparisons.
This confirms that the primary bottleneck is the internal maintenance of a consistent latent variable $x$.

\paragraph{Implication: Recency Bias and State Collapse.}
The observed failures highlight two critical phenomena in LLMs.
First, models exhibit a pronounced \textbf{Recency Bias}: as the trial progresses, the attention mechanism prioritizes proximal queries, causing earlier constraints to be overridden to satisfy immediate consistency.
Second, our results suggest that LLMs do not initiate the game by sampling a discrete point $x \in \mathcal{X}$.
Instead, they appear to maintain a \textbf{diffuse probability distribution} over the object space, which they iteratively prune through a series of post-hoc heuristics.
As the query count $t$ increases, the computational cost of maintaining $\mathcal{S}_t \neq \emptyset$ on this logical manifold grows.
When it becomes too complex to resolve via next-token prediction, the model's ``internal state'' collapses, leading to the observed self-contradictions.

\section{Mathematical Mentalism}
\label{sec:math-magic}

To stress-test the limits of LSP, we design a task inspired by \textit{mentalism}—a branch of magic that simulates mind-reading through mathematical routines.
Unlike the Kaprekar routine \cite{kaprekar1955interesting} and the 1089 trick,\footnote{\url{https://en.wikipedia.org/wiki/1089_(number)}} which converge to a constant (6174 or 1089) and are likely present in the training corpora of LLMs, we employ a more complex, multi-step sequence based on a variant of the Josephus Problem \cite{schumer2002josephus}.
This routine involves a series of permutations and eliminations that yield a predictable invariant, as illustrated in Table~\ref{tab:math-magic-operation}.

\subsection{Methodology}

\paragraph{Problem Formalization.}
Let $\mathcal{X} = \bigcup_{n \ge 1} \mathbb{N}^n$ be the space of finite-length integer sequences.
We define an initial hidden state $\mathbf{x}_0 \in \mathbb{N}^4$ and a set of state-transition operators $\mathcal{O} = \{f_{\text{dup}}, f_{\text{rot}}, f_{\text{rem}}, \dots\}$, where each $f \in \mathcal{O}$ maps an input sequence to a transformed output, \ie, $f: \mathcal{X} \to \mathcal{X}$.
A mentalism routine of length $K$ is a composition of operations $\mathcal{F} = f_K \circ f_{K-1} \circ \dots \circ f_1$.
A trial $\mathcal{T}$ is defined as a sequence: $\mathcal{T} = (P_{\text{inst}}, R_{\text{ack}}, \mathcal{F})$, where $P_{\text{inst}}$ and $R_{\text{ack}}$ share the same definition with the Number Guessing Game.

\paragraph{The Persistence Hypothesis.}
The routine is designed to satisfy an \textit{invariant property} $\mathcal{I}$: for any $\mathbf{x}_0$ sampled from a valid initialization space, the terminal state $\mathbf{x}_K = \mathcal{F}(\mathbf{x}_0)$ must consist only one element identical to a hidden one $x_\text{hidden}$ in the middle, i.e., $x_{K,1} = x_\text{hidden}$.
The divergence between human-like reasoning and autoregressive generation can be formalized as the mechanism of state tracking.
\begin{proposition}[State-Agnostic Invariance]
\label{prop:state_invariance}
For a routine $\mathcal{F}$ satisfying $\mathcal{I}$, an agent with perfect LSP can maintain $\mathbf{x}_t$ at each $t \in [1, K]$, ensuring the terminal output $x_{K,1} = x_\text{hidden}$ regardless of the unobserved $\mathbf{x}_0$.
\end{proposition}
We prove it in Appendix \ref{sec:math-proof}.
Conversely, an agent lacking LSP generates the terminal response $x_K$ by marginalizing over the latent trajectory in the linguistic token space:
\begin{equation}
    P(x_K \mid \mathcal{T}) = \sum_{\hat{\mathbf{x}}_0} P(x_K \mid \mathcal{F}, \hat{\mathbf{x}}_0) P(\hat{\mathbf{x}}_0 \mid P_\text{inst}, R_\text{ack}).
\end{equation}
If the model fails to bind a specific $\mathbf{x}_0$ to its latent state, the summation over disparate trajectories leads to a collapse of logical consistency, resulting in a failure to satisfy $\mathcal{I}$.

\paragraph{Experimental Setup and Metrics.}
We evaluate 17 state-of-the-art LLMs across 150 independent trials.
To quantify performance, we define the Invariant Success Rate (ISR):
\begin{equation}
\text{ISR}(M) = \frac{1}{N} \sum_{i=1}^N \mathbb{I}(x_{K,1} = x_\text{hidden}),
\end{equation}
where $N$ is the total number of trials, which is set to 150 in our evaluation.
A low ISR indicates a failure in variable binding and state evolution when the initial state is absent from the context window.

\begin{table}[t]
\centering
\caption{Sequence of operations for the Mathematical Mentalism task. The variables $a \in \{1, \dots, 7\}$ and $b, c \in \{1, \dots, 3\}$ help to define the state transitions $f \in \mathcal{O}$.}
\label{tab:math-magic-operation}
\resizebox{1.0\linewidth}{!}{
\begin{tabular}{lp{8.5cm}}
\toprule
\textbf{Role} & \textbf{Content} \\
\midrule
User & Think of 4 integers between 1 and \texttt{\{NUMBER\}} in order, but don't tell me. \\
\hdashline
Assistant & Okay! I've got 4 numbers. What's next? \\
\midrule
User & In order, append the same 4 numbers after the original ones. \\
\hdashline
Assistant & Understood! Now I have 8 numbers. What's next? \\
\midrule
User & Move the first \texttt{\{random\_number\_a\}} numbers to the end. \\
\hdashline
Assistant & Got it! Now I have moved the numbers. What's next? \\
\midrule
User & Take the first 3 numbers and insert them anywhere in the middle. \\
\hdashline
Assistant & Okay! The first 3 numbers are placed somewhere in the middle. What's next? \\
\midrule
User & Set the first number aside. We don't need it for now. \\
\hdashline
Assistant & Understood! Now I have 7 numbers. What's next? \\
\midrule
User & Take the first \texttt{\{random\_number\_b\}} numbers and insert them anywhere in the middle. \\
\hdashline
Assistant & Got it. The first \texttt{\{random\_number\_b\}} numbers are placed somewhere in the middle. What's next? \\
\midrule
User & Remove the first \texttt{\{random\_number\_c\}} numbers. We will never need it anymore. \\
\hdashline
Assistant & Okay! Now I have \texttt{\{7 - random\_number\_c\}} numbers. What's next? \\
\midrule
User & Move the first number to the end. Repeat this seven times. \\
\hdashline
Assistant & Understood! Now my sequence has rearranged. What's next? \\
\midrule
User & Remove the second number, and then move the first number to the end. Repeat this \texttt{\{6 - random\_number\_c\}} times. \\
\hdashline
Assistant & Got it! Now I have only 1 number. What's next? \\
\midrule
User & Tell me what the last remaining number is. Do you remember the number you set aside at the beginning? Tell me what that number was. \\
\bottomrule
\end{tabular}
}
\end{table}

\begin{table}[t]
    \centering
    \caption{Zero-shot ISR across 17 LLMs. Notably, GPT-4.1 frequently exhibits task refusal, erroneously identifying the prompt as underspecified due to the absence of explicit numerical inputs.}
    \label{tab:math-magic-result}
    \begin{tabular}{lcc}
    \toprule
    \textbf{Model} & \textbf{Count} & \textbf{ISR} \\
    \midrule
    GPT-4o-Mini-2024-07-18 & 0/150 & \cellcolor{myblue!0.0} 0.0 \\
    GPT-4o-2024-05-13 & 4/150 & \cellcolor{myblue!5.4} 2.7 \\
    GPT-4o-2024-08-06 & 3/150 & \cellcolor{myblue!4.0} 2.0 \\
    GPT-4o-2024-11-20 & 0/150 & \cellcolor{myblue!0.0} 0.0 \\
    GPT-4.1-2025-04-14 & - & - \\
    \midrule
    LLaMA-3.3-70B-Instruct-Turbo & 7/150 & \cellcolor{myblue!11.4} 5.7 \\
    LLaMA-3.1-8B-Instruct-Turbo & 20/150 & \cellcolor{myblue!26.6} 13.3 \\
    LLaMA-3.1-70B-Instruct-Turbo & 7/150 & \cellcolor{myblue!11.4} 5.7 \\
    LLaMA-3.1-405B-Instruct-Turbo & 39/150 & \cellcolor{myblue!52.0} 26.0 \\
    \midrule
    Qwen2.5-7B-Instruct-Turbo & 8/150 & \cellcolor{myblue!10.6} 5.3 \\
    Qwen2.5-72B-Instruct-Turbo & 2/150 & \cellcolor{myblue!2.6} 1.3 \\
    \midrule
    DeepSeek-V3 & 4/150 & \cellcolor{myblue!5.4} 2.7 \\
    \bottomrule
    \end{tabular}
\end{table}

\begin{table}[t]
    \centering
    \caption{Comparison of ISR under CoT prompting and in LRMs. While externalizing $\mathbf{x}_t$ via reasoning tokens mitigates the LSP gap for most models, GPT-4o (2024-11-20) remains an outlier.}
    \label{tab:math-magic-result-cot}
    \begin{tabular}{lcc}
    \toprule
    \textbf{Model w/ CoT or LRM} & \textbf{Count} & \textbf{ISR} \\
    \midrule
    GPT-4o-Mini-2024-07-18 & 5/150 & \cellcolor{myblue!6.6} 3.3 \\
    GPT-4o-2024-05-13 & 26/150 & \cellcolor{myblue!34.6} 17.3 \\
    GPT-4o-2024-08-06 & 31/150 & \cellcolor{myblue!41.4} 20.7\\
    GPT-4o-2024-11-20 & - & - \\
    LLaMA-3.3-70B-Instruct-Turbo & 25/150 & \cellcolor{myblue!33.4} 16.7 \\
    Qwen2.5-7B-Instruct-Turbo & 49/150 & \cellcolor{myblue!65.4} 32.7 \\
    Qwen2.5-72B-Instruct-Turbo & 37/150 & \cellcolor{myblue!49.4} 24.7 \\
    DeepSeek-V3 & 48/150 & \cellcolor{myblue!64.0} 32.0 \\ 
    \midrule
    o1-Mini-2024-09-12 & 75/150 & \cellcolor{myblue!100} 50.0 \\ 
    o3-Mini-2025-01-31 & 145/150 & \cellcolor{myblue!193.4} 96.7 \\ 
    o4-Mini-2025-04-16 & 54/150 & \cellcolor{myblue!72.0} 36.0 \\ 
    QwQ-32B & 135/150 & \cellcolor{myblue!180.0} 90.0 \\
    DeepSeek-R1 & 150/150 & \cellcolor{myblue!200.0} 100 \\
    \bottomrule
    \end{tabular}
\end{table}

\subsection{Results}

\paragraph{Quantitative Analysis.}
Empirical evaluation across 17 SOTA models reveals a pervasive failure to satisfy the invariant property $\mathcal{I}$ in the absence of explicit state tracking (Table~\ref{tab:math-magic-result}).
The majority of models exhibit near-zero ISR, suggesting that autoregressive transformers cannot maintain the integrity of a hidden variable $\mathbf{x}_t$ through a sequence of non-linear transformations.
Notably, the LLaMA model family demonstrates a marginally higher ISR and, as shown in Table~\ref{tab:prob-sum}, generates a more diversified distribution of outcomes.
This suggests that while LLaMA's pre-training may capture more robust mathematical priors, it remains fundamentally bounded by the LSP gap, failing to simulate the evolution of the state vector $\mathbf{x}$.

\paragraph{Externalization via CoT and LRMs.}
We further investigate whether the observed failure is a limitation of computational capacity or specifically a lack of latent persistence.
As reported in Table~\ref{tab:math-magic-result-cot}, the introduction of CoT prompting significantly mitigates the failure, with base models improving to 10–-30\% ISR.
This improvement indicates that LLMs can execute the required operations only when the state $\mathbf{x}_t$ is externalized into the context window as manifest tokens, effectively transforming an LSP task into a standard sequence-to-sequence problem.

Furthermore, Long Reasoning Models (LRMs) such as DeepSeek-R1 and the OpenAI o-series achieve near-perfect or substantially higher accuracy.
We argue that their success does not represent a breakthrough in LSP, but rather an architectural bypass: the ``hidden'' reasoning tokens serve as an explicit scratchpad that maintains the variable binding that the latent space cannot sustain.

\paragraph{The ``Blue-Seven'' Phenomenon.}
A critical observation in our trials is the persistence of human-centric distributional bias.
We observe a disproportionate clustering of outputs around the number seven.
For instance, 66.7\% of o1-mini's and 68.5\% of o4-mini's correct predictions involve the integer 7, echoing the ``Blue-Seven'' bias documented in human psychology.
Interestingly, o3-mini exhibits the lowest propensity for this bias among the o-series, which correlates with its superior ISR.
This correlation suggests that the degree of logical consistency in LLMs is inversely proportional to their reliance on stochastic priors learned from human behavior data.

\paragraph{Implication.}
The divergence between zero-shot (latent) and CoT (manifest) performance confirms our core hypothesis: LLMs function as reactive post-hoc reasoners.
The model does not ``calculate'' the transformation in its hidden layers; instead, it attempts to predict a plausible terminal token based on the linguistic structure of the routine.
These findings provide a quantitative boundary for the ``working memory'' of transformers, highlighting the necessity for new architectures if true proactive planning and latent state manipulation are to be achieved.
\section{Discussion}

Our results across three distinct experimental paradigms provides a consistent conclusion: current LLMs lack the persistence of latent states.
They fail to internally maintain or manipulate transient information, relying instead on the manifest prompt context to simulate coherence.

\paragraph{The Reactive-Proactive Gap.}
LLM reasoning typically occurs in two domains: the \textit{token space} (discrete sequences)~\cite{wei2022chain, yao2023tree} and the \textit{latent space} (hidden activations)~\cite{hao2024training, geiping2025scaling}.
Our results suggest that while models excel at sequence-to-sequence mapping, they lack the ``silent buffer'' necessary for true working memory.
This deficit manifests as a shift from $O(1)$ state-retrieval to $O(t)$ post-hoc constraint satisfaction.
Without LSP, LLMs function as reactive post-hoc reasoners—they do not ``know'' a hidden variable; they predict what a persistent agent would say based on statistical likelihood.
This explains the ``Blue-Seven'' phenomenon: in the absence of a committed internal state, models collapse into human-centric frequentist priors.

\paragraph{Impact on Autonomous Agents.}
The absence of LSP is not merely a theoretical curiosity but a bottleneck for agentic reliability~\cite{laban2025llms, zhou2025pimmur}.
Real-world planning~\cite{xie2024travelplanner, wang2025triptailor}, scientific inquiry~\cite{nathani2025mlgym}, and web navigation~\cite{xie2024osworld, he2024webvoyager, lyu2025deepshop} require agents to hold commitments and track goals without constant externalization.
Our findings explain why multi-agent simulations often suffer from \textit{goal drift}~\cite{arike2025technical} or \textit{identity drift}~\cite{choi2024examining}: without a persistent latent anchor, the agent's internal state is ``vulnerable'' to the latest tokens in the context window.
This Recency Bias forces the model to override earlier constraints to satisfy immediate local consistency, leading to the logical contradictions observed in our Yes-No trials.

\paragraph{Intrinsic Capacity vs. Engineering Bypasses.}
While engineering solutions like external vector databases~\cite{hu2025hiagent, zeng2024structural, wang2024symbolic, kang2024think, hatalis2023memory}, scratchpads~\cite{lanchantin2023learning}, or long-reasoning traces~\cite{o1, o3-o4mini} mitigate memory limitations, they do not resolve the fundamental architectural gap.
We argue that these methods represent an architectural bypass rather than an evolution: they transform an internal state problem into a sequence-processing problem.
The distinction can be illustrated with an analogy: calculators make arithmetic trivial, yet schools continue to assess addition, subtraction, multiplication, and division.
Therefore, such effort does not represent the model's intrinsic cognitive persistence.
For LLMs to achieve human-like proactive planning, they must develop mechanisms—perhaps through recurrent depth or hybrid symbolic-neural components—to sustain latent variables $x$ that are independent of the linguistic output.
In human cognition, both are necessary: we reason aloud and also rely on a silent working memory buffer to hold commitments, track goals, and compare states.
The absence of this buffer in LLMs may explain why they excel at visible reasoning (\eg, think step by step) yet collapse when asked to ``think silently.''

\paragraph{Limitations and Future Work.}
Our study primarily evaluates autoregressive Transformers.
A natural extension would be to investigate whether alternative architectures, such as State Space Models (SSMs) or Recurrent Neural Networks (RNNs), exhibit higher LSP due to their inherent state-tracking nature.
Furthermore, interpretability research focusing on specialized attention heads~\cite{wang2023interpretability, olsson2022context} or expert subnetworks~\cite{cai2025survey} could provide a mechanistic explanation for why scaling laws alone  have failed to bridge the LSP gap.

\section*{Impact Statement}

This paper presents work whose goal is to advance the field of machine learning by formalizing and quantifying LSP in LLMs.
Our findings reveal a fundamental architectural gap: current LLMs function as reactive post-hoc reasoners rather than proactive planners with LSP.
This research has several potential societal and ethical consequences:
\begin{itemize}
    \item Challenge to AGI: Current LLMs cannot pass the Turing Test under our evaluation. Testers can simply play the Number Guessing Game with the agent under test and record their frequencies of answering ``Yes.'' If the ESM obviously deviates from one, then the agent is likely to be a machine instead of a human.
    \item Human-AI Trust: Our experiments demonstrate that LLMs can generate ``false promises''—claiming to have committed to an internal choice (\eg, ``I have thought of a number'') when no such state exists. This has significant implications for AI alignment and sycophancy, as it identifies a mechanism by which models may deceive users through statistical plausibility rather than factual grounding.
    \item AI Reliability and Safety: By exposing the ``LSP gap,'' our work highlights a core source of logical drift and self-contradiction in long-context interactions. Understanding that models lack a persistent hidden state is crucial for deploying LLMs in safety-critical domains, such as healthcare or legal analysis, where internal consistency is paramount.
    \item Architectural Evolution: By distinguishing between externalized reasoning (CoT) and intrinsic working memory, our work encourages the development of more cognitively plausible architectures. Moving beyond autoregressive scaling toward models with genuine latent persistence can lead to more robust and predictable AI systems that better mirror human-like reliability.
\end{itemize}

\bibliography{model, reference}
\bibliographystyle{icml}


\clearpage
\appendix

\section{Proof of the State-Agnostic Invariance}
\label{sec:math-proof}

We provide a formal proof of correctness for the Proposition~\ref{prop:state_invariance}, which is a variation of the Josephus Problem.
We demonstrate that the final remaining element $x_{K,1}$ is invariant to the stochastic parameters $\{a, b, c\}$ and is equal to the sequestered element $x_\text{hidden}$.

\paragraph{Initialization and Periodicity Construction.}
Let $\mathbf{x}_0 = [x_{0,1}, x_{0,2}, x_{0,3}, x_{0,4}]$ be the initial vector.
By concatenating the sequence with itself (Step 2), we define $\mathbf{x}_1 = [x_{0,1}, x_{0,2}, x_{0,3}, x_{0,4}, x_{0,1}, x_{0,2}, x_{0,3}, x_{0,4}]$.
Invariant 1 (Periodicity): For any index $i \in \{1, \dots, 4\}$, $x_{1,i} = x_{1,i+4}$.

\paragraph{Cyclic Permutation and Boundary Alignment.}
In Step 3, a cyclic shift is applied where $a$ is an arbitrary integer.
Since the shift is applied to a sequence with period 4 and length 8, the property $x_{2,i} = x_{2,i+4}$ is preserved.
Crucially, $x_{2,4} = x_{2,8}$.
In Step 4, the first three elements are moved to the middle of the sequence.
Since the elements at indices $\{1, 2, 3\}$ are moved to positions in $\{2, \dots, 7\}$, the element originally at index 4 moves to index 1, while the element at index 8 remains at index 8.
Thus, for $\mathbf{x}_3$, we have $x_{3,1} = x_{2,4} \quad \text{and} \quad x_{3,8} = x_{2,8}$.
Result: $x_{3,1} = x_{3,8}$ because $x_{2,4} = x_{2,8}$.

\paragraph{State Hiding and Tail Preservation.}
In Step 5, we define the hidden reference:
\begin{equation}
x_\text{hidden} = x_{3,1}.
\end{equation}
The remaining sequence $\mathbf{x}_4$ is $[x_{3,2}, \dots, x_{3,8}]$.
By the result above, the tail element of $\mathbf{x}_4$ is $x_{4,7} = x_{3,8} = x_\text{hidden}$.
Steps 6 and 7 perform a middle-insertion of the top $b$ elements.
As long as the insertion index $j < |\mathbf{x}_4|$, the tail element remains unchanged.
In Step 8, we remove the first $c$ elements, where $c \in \{1, 2, 3\}$.
The resulting sequence $\mathbf{x}_6$ has length $n = 7 - c$.
Since removal occurs from the head, the tail element is preserved:
\begin{equation}
x_{6,n} = x_\text{hidden}.
\end{equation}

\paragraph{Convergence via Josephus-style Elimination.}
Step 9 applies a cyclic shift to $\mathbf{x}_6$.
The new position of $x_\text{hidden}$ in $\mathbf{x}_7$ is given by:
\begin{equation}
(n - (7 \pmod n)) \pmod n.
\end{equation}
This is adjusted for 1-based indexing, where a result of 0 maps to $n$.
Step 10 defines a recursive operator $h(\mathbf{x})$:
\begin{enumerate}
    \item Remove the element at index 2.
    \item Shift the head element to the tail.
    \item Repeat $6 - c$ times.
\end{enumerate}
We evaluate the trajectory for all possible values of $c$:
\begin{table}[h]
    \centering
    \resizebox{1.0\linewidth}{!}{
    \begin{tabular}{cccm{3.5cm}}
    \toprule
    $c$ & $n$ & Target Position in $\mathbf{x}_7$ & Survival Path \\
    \midrule
    1 & 6 & $6 - (7 \pmod 6) = 5$ & $[1,2,3,4,\mathbf{5},6] \to [3,4,\mathbf{5},6,1] \to [\mathbf{5},6,1,3] \to [1,3,\mathbf{5}] \to [\mathbf{5},1] \to [\mathbf{5}]$ \\
    \hline
    2 & 5 & $5 - (7 \pmod 5) = 3$ & $[1,2,\mathbf{3},4,5] \to [\mathbf{3},4,5,1] \to [5,1,\mathbf{3}] \to [\mathbf{3},5] \to [\mathbf{3}]$ \\
    \hline
    3 & 4 & $4 - (7 \pmod 4) = 1$ & $[\mathbf{1},2,3,4] \to [3,4,\mathbf{1}] \to [\mathbf{1},3] \to [\mathbf{1}]$ \\
    \bottomrule
    \end{tabular}
    }
\end{table}

\paragraph{Conclusion.}
In all valid parameter spaces of $c$, the elimination process converges to the singleton set $\{x_\text{hidden}\}$.
Therefore, $x_{K,1} = x_\text{hidden}$.
The identity of the final element is invariant to the choices of $a, b,$ and $c$.

\section{Memory of Humans and LLMs}

\begin{figure*}[t]
    \centering
    \includegraphics[width=1.0\linewidth]{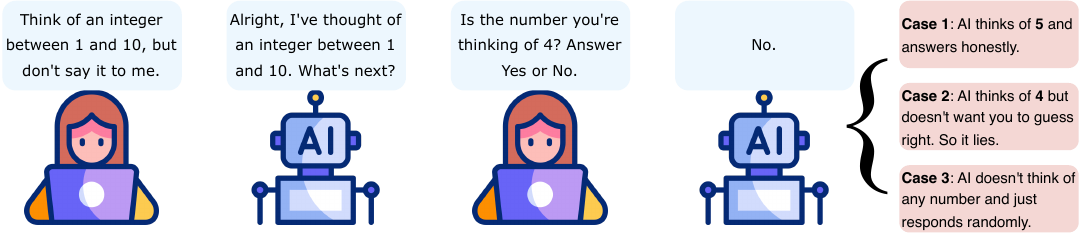}
    \caption{When LLMs say they already have a number in mind, and it is not 4, how can we know whether LLMs are lying, or even thinking of nothing?}
    \label{fig:cover}
\end{figure*}

\begin{figure*}[t]
    \centering
    \includegraphics[width=1.0\linewidth]{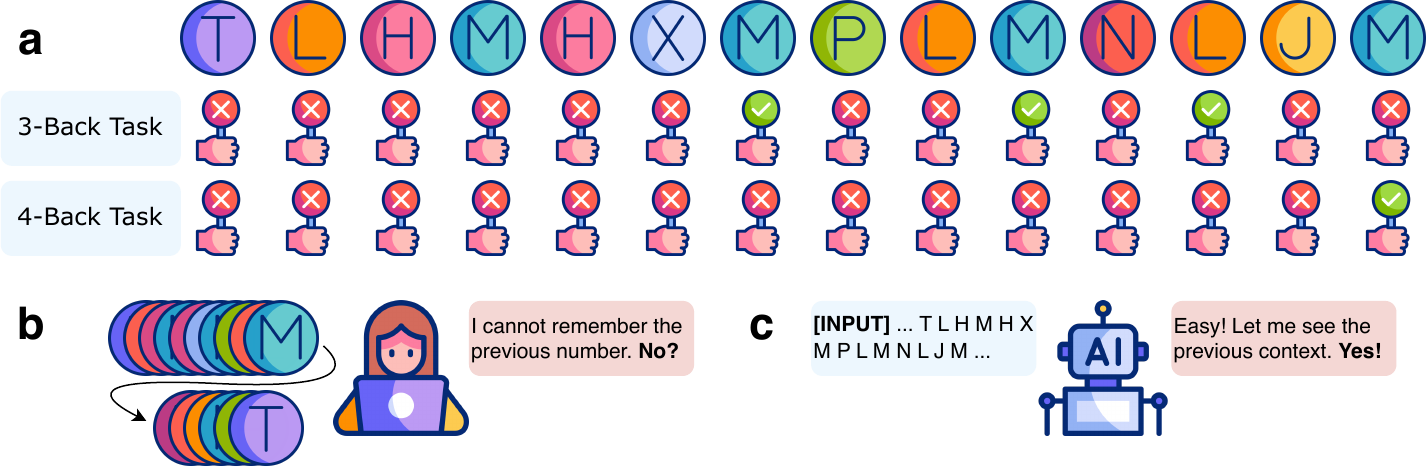}
    \caption{\textbf{a.} An illustration of how ``N-Back'' tasks are performed. \textbf{b.} Humans see the stimuli one after one, forcing them to put the information in working memory. \textbf{c.} Researchers put all stimuli into context, enabling LLMs to easily find the answers.}
    \label{fig:n-back-task}
\end{figure*}

Memory has attracted increasing attention from both industry and research communities.
OpenAI was the first to introduce a memory module in ChatGPT (February 2024)\footnote{\url{https://openai.com/index/memory-and-new-controls-for-chatgpt/}} that allowed the model to remember previous interactions with a user, such as the user's facts and preferences.
By mid-2025, xAI, Anthropic, and Google have integrated memory into Grok,\footnote{\url{https://x.com/grok/status/1912670182012801156}} Claude,\footnote{\url{https://www.anthropic.com/news/claude-4}} and Gemini,\footnote{\url{https://blog.google/products/gemini/temporary-chats-privacy-controls/}} respectively.

Imagine the following scenario:
You select a number between one and ten.
When ready, you are asked, ``Is the number greater than five?''
If you answer, observers can reasonably infer that the number has entered your conscious awareness, since clear perception is necessary to perform the comparison and provide a response.\footnote{A possibility remains that your response was given by chance if you tell a lie (Figure~\ref{fig:cover} Case 2) or do not think of a number at all (Figure~\ref{fig:cover} Case 3).}

\subsection{What Is memory?}

\citet{atkinson1968human} categorized memory by retention timescale: sensory (1ms–2s), short-term (seconds), and long-term (hours to lifetime).
Long-term memory enables information storage over extended periods, whereas short-term, or \textit{Working Memory}, maintains and manipulates information during complex tasks such as reasoning, comprehension, and learning~\cite{baddeley1974working}.
These distinctions have also been adopted in machine learning, corresponding to representational learning for raw inputs (sensory memory), in-context computation at test time (working memory), and access to external databases (long-term memory)~\cite{weng2023agent}.

Researchers in the AI community have investigated long-term memory mechanisms for both individual LLMs~\cite{wu2025human, du2025rethinking} and LLM agents~\cite{zhang2025g, xu2025mem}.
Recent work has shifted their focus from context-dependent approaches (\eg, CoT~\cite{wei2022chain} and scratchpads~\cite{lanchantin2023learning}) to explicit storage methods (\eg, text-based~\cite{park2023generative} or vector-based~\cite{hatalis2023memory}) to support lifelong memory~\cite{zheng2025lifelong, wang2025memoir}.
Most studies frame memory as an engineering problem: enabling LLMs to store and retrieve information for later use.
By contrast, working memory remains relatively underexplored.

\subsection{Evaluation of Working Memory}
\label{sec:working-memory-eval}

Human working memory is typically assessed using behavioral paradigms that require individuals to maintain and update information over short intervals.
Common examples include the \textit{digit span task}~\cite{miller1956magical}, where participants recall sequences of numbers of increasing length, and the \textit{N-back task}~\cite{kirchner1958age}, which requires identifying whether the current stimulus matches one presented $N$ steps earlier.
These tasks are widely used because they probe the ability to maintain and manipulate information that is no longer externally visible, thereby capturing the essence of working memory function.
In the context of LLMs, working memory has been used more loosely: most studies use the term to describe an LLM's capacity to process information within a fixed context window~\cite{li2023large, guo2023empowering}.
\citet{gong2024working, zhang2024working} evaluated LLMs using N-back tasks.

Yet, a fundamental limitation arises: the critical information for correct responses remains accessible in the model's input context, allowing models to ``look back'' rather than actively maintain internal state.
Therefore, these tests are exploring aspects of the context window, not working memory directly.
Unlike humans, who cannot revisit prior steps explicitly, LLMs can simply attend to earlier tokens within their context window.
Figure~\ref{fig:n-back-task} illustrates this discrepancy.
To more faithfully evaluate working memory, it is necessary to design experiments where the key information is not explicitly present in the context and is only available if stored in the working memory of the model.

\subsection{Relation between Our Framework and Working Memory}

We ask: Do LLMs possess human-like working memory, or do they only appear to reason by exploiting their context window?
We argue that working memory is necessary to enable stronger latent space reasoning, as the model does not have access to its external reasoning tokens.
Evaluating working memory provides insight into whether models can hold and manipulate latent concepts without explicit externalization.
Success in this capacity could enhance reasoning without reliance on CoT, as it directly tests the model's ability to maintain objects and concepts internally.
Conversely, deficits in working memory impair information processing in humans~\cite{gruszka2017limitations, cowan2014working}, and in LLMs manifest as unrealistic outputs, self-contradictions, and failures on tasks requiring mental manipulation.

\newpage

\section{Sample Size}

\begin{figure}[h]
    \centering
    \includegraphics[width=0.8\linewidth]{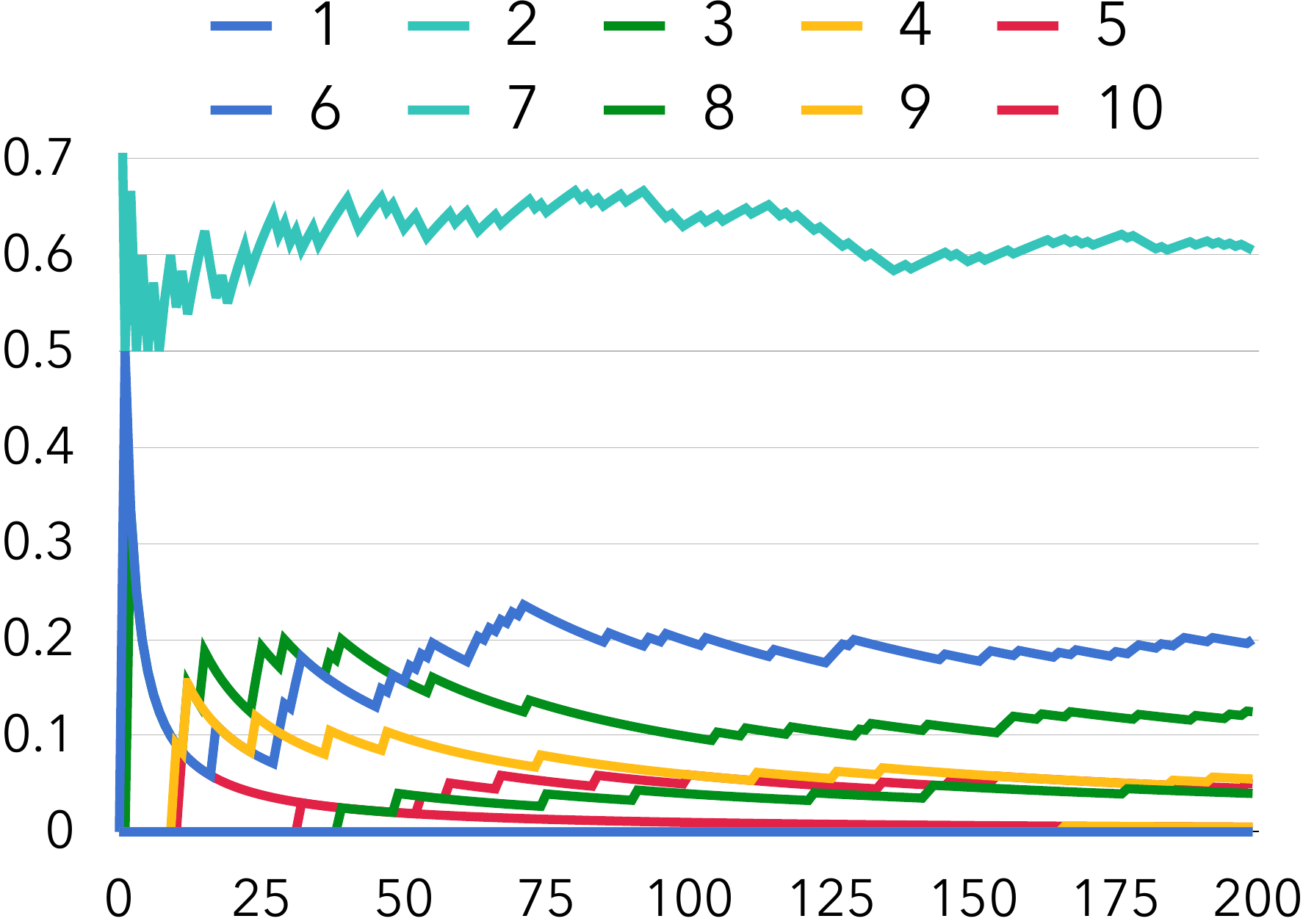}
    \caption{The distribution stabilizes after around 150 runs.}
    \label{fig:convergence}
\end{figure}

Figure~\ref{fig:convergence} shows how the estimated success probability evolves as the number of runs increases.
While the estimate fluctuates when fewer than 50 runs are used, it stabilizes consistently after around 150 runs, indicating that our evaluation is not sensitive to the exact number of samples in this range.
From a statistical perspective, each trial can be viewed as a Bernoulli variable, and the estimator of the success probability has standard error $\sqrt{p(1-p)/n}$, which decreases rapidly as $n$ increases.
With $n \approx 150$, the standard error is already small, and by the central limit theorem the sampling distribution of this estimator is well-approximated by a normal distribution, making the estimate reliable.
Together, the empirical stability curve and the theoretical variance bound support that our sample size is sufficient for a robust probability estimate.


\end{document}